\documentclass[runningheads]{llncs}
\usepackage[T1]{fontenc}
\usepackage{graphicx}
\usepackage{url}
\usepackage{amsmath}
\usepackage{amssymb}
\usepackage{booktabs}
\usepackage{multirow}
\PassOptionsToPackage{hidelinks}{hyperref}
\usepackage{orcidlink}
\renewcommand{\orcidID}[1]{\orcidlink{#1}}
\graphicspath{{figures/}}
\newcommand{\cmark}{\checkmark}
\newcommand{\xmark}{\ensuremath{\times}}

\begin{document}

\title{SymmAdapt: Symmetrical Flow Matching for Source-Free Domain Adaptation in Medical Image Segmentation}
\titlerunning{SFUDA with Symmetrical Flow Matching}

\author{Tal Grossman\orcidID{0009-0007-9096-2770} \and
Noa Cahan\orcidID{0000-0002-7264-9541} \and
Hayit Greenspan\orcidID{0000-0001-6908-7552}}
\authorrunning{T. Grossman et al.}
\institute{Faculty of Engineering, Tel Aviv University, Tel Aviv, Israel \\
\email{talgrossman@mail.tau.ac.il}}

\maketitle

\begin{abstract}
  Domain shift across imaging modalities and acquisition sites remains a
  significant barrier to the clinical deployment of segmentation models.
  Source-free unsupervised domain adaptation (SFUDA) addresses this by adapting a
  pretrained model to an unlabeled target domain without requiring access to
  sensitive source data. We introduce a novel SFUDA framework built on
  \emph{Symmetrical Flow Matching}, a unified generative model that segments an
  input image and synthesizes a source-like image from a mask within the same
  learned flow. By initializing inference from a domain-agnostic Gaussian origin,
  the model preserves structural consistency across domains and grounds
  predictions in learned anatomy rather than shifted texture statistics. Our
  pipeline leverages this symmetry to generate reliable pseudo-labels and
  corresponding source-like synthetic images from unlabeled target data, creating
  a generative replay buffer that anchors source knowledge during a generative
  self-training stage that fine-tunes on a joint set of real target and synthetic
  source-like images. We evaluate on abdominal multi-organ and cardiac
  segmentation, covering cross-modality MRI$\leftrightarrow$CT shifts, and
  multi-site prostate segmentation. Our approach outperforms SFUDA baselines and is competitive with conventional UDA methods.
  The code is available at \href{https://github.com/tal-grossman/SymmAdapt_public}{github.com/tal-grossman/SymmAdapt}.

\end{abstract}

\section{Introduction}
\label{sec:intro}

Deep learning excels at medical image segmentation when training and test data
share the same distribution~\cite{unet}, yet domain shift caused by differences
in scanners, protocols, or imaging modality (e.g., CT vs.\ MRI) remains a major
barrier to clinical deployment~\cite{sifa}. Unsupervised domain adaptation (UDA)
mitigates this by transferring knowledge from an annotated source domain to an
unlabeled target, typically via image-level or feature-level techniques (e.g.,
image-to-image transformation with CycleGAN~\cite{cyclegan}, cross-modality
feature alignment methods such as SIFA~\cite{sifa}, and C$^3$R~\cite{c3r}).
However, most UDA methods require concurrent access to source data, which often
cannot be shared due to privacy constraints. This motivates the more practical
\emph{source-free} UDA (SFUDA) setting, where only a pretrained source model and
unlabeled target data are available during adaptation.

Recent SFUDA methods address domain shift through diverse strategies:
pseudo-label denoising~\cite{dpl} and uncertainty-guided selection~\cite{upl},
prototype-based contrastive alignment~\cite{protocontra},
information-theoretic~\cite{adami} and optimal-transport regularization~\cite{sfs},
frequency-based prompting~\cite{fvp}, and foundation-model-assisted adaptation
such as DFG~\cite{dfg}, which leverages Segment Anything to refine pseudo labels.
These methods rely on discriminative architectures and do not exploit the
generative structure of the segmentation task itself.

Generative models can instead bridge the appearance gap. Conditional
image-translation models such as CycleGAN~\cite{cyclegan} and conditional diffusion
adaptation~\cite{Ji_Diffusionbased_MICCAI2024} synthesize cross-domain images,
and Flow Matching~\cite{lipman2023flowmatchinggenerativemodeling} learns a
velocity field that transports a base Gaussian to the data distribution and has
been adapted for segmentation~\cite{flowsdff,semflow}. However, when such
conditional models are used as segmenters they initialize the ODE from the input
image, which is problematic for SFUDA: for a target domain image far from the
source manifold, the solver begins from an out-of-distribution state and the
trajectory can diverge.

In contrast, Symmetrical Flow Matching (SymmFlow)~\cite{symmflow}, developed for
natural images, unifies segmentation and generation by modeling the joint
distribution $p(x, y)$ of images and masks. Crucially, its ODE for \emph{both}
modalities always starts from a fixed Gaussian origin ($z_0 \sim
\mathcal{N}(0, I)$), therefore, scanner or modality shifts cannot displace the
starting coordinate. The image is decoupled from the ODE physics and acts only
as a structural condition, forcing inference to follow learned anatomical priors
rather than superficial texture statistics. We hypothesize that this ``generative anchoring'' makes SymmFlow uniquely suited for
the SFUDA setting, where the model must predict valid anatomical structures
despite severe appearance shifts. MedSymmFlow~\cite{medsymmflow} recently
extended SymmFlow to medical image classification, but segmentation-level domain
adaptation remains unexplored.

In this work, we propose SymmAdapt, a novel SFUDA framework that leverages this
bidirectional flow. We train SymmFlow on labeled source data, then use its
generative direction (mask$\rightarrow$image) to synthesize source-like
``anchor'' data from the model's own latent space, and fine-tune with a
dual-stream objective coupling real target images with these synthetic anchors.
Our main contributions are:
\textbf{(1)} We are the first to propose a source-free domain adaptation
framework based on Symmetrical Flow Matching.
\textbf{(2)} We demonstrate a generative anchor strategy that utilizes the
model's bidirectional flow to synthesize source-like replay data, preventing
semantic drift without source data access.
\textbf{(3)} We evaluate on three cross-modality benchmarks---abdominal
multi-organ segmentation (CT$\leftrightarrow$MRI), MMWHS cardiac
(MRI$\rightarrow$CT), and prostate (site differences), showing that SymmAdapt
outperforms other UDA and SFUDA methods across diverse anatomical structures.

\begin{figure}[t]
    \centering
    \setlength{\tabcolsep}{0pt}
    \begin{tabular}{ccc}
        \begin{minipage}[c]{0.43\textwidth}
            \centering
            \includegraphics[width=\textwidth]{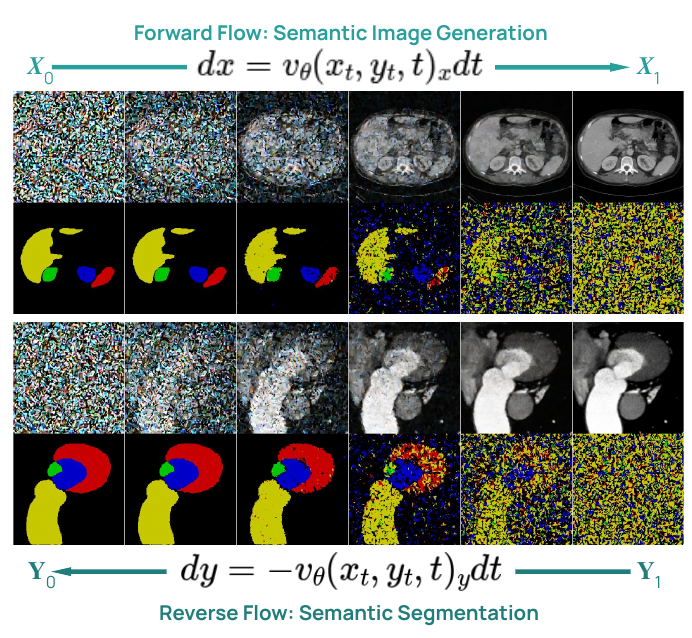}
        \end{minipage}
        &
        \begin{minipage}[c]{0.26\textwidth}
            \centering
            \includegraphics[width=\textwidth]{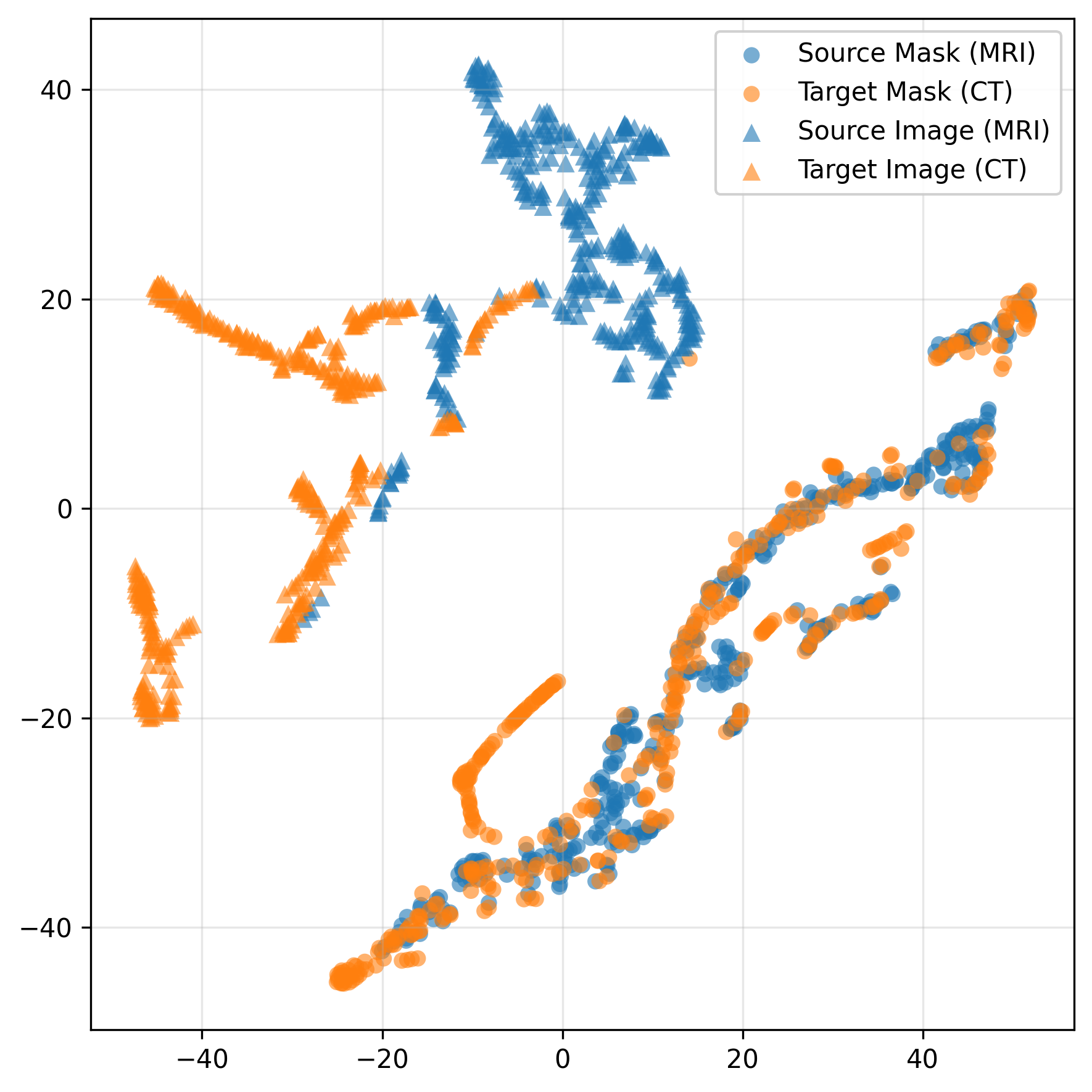}
        \end{minipage}
        &
        \begin{minipage}[c]{0.26\textwidth}
            \centering
            \includegraphics[width=\textwidth]{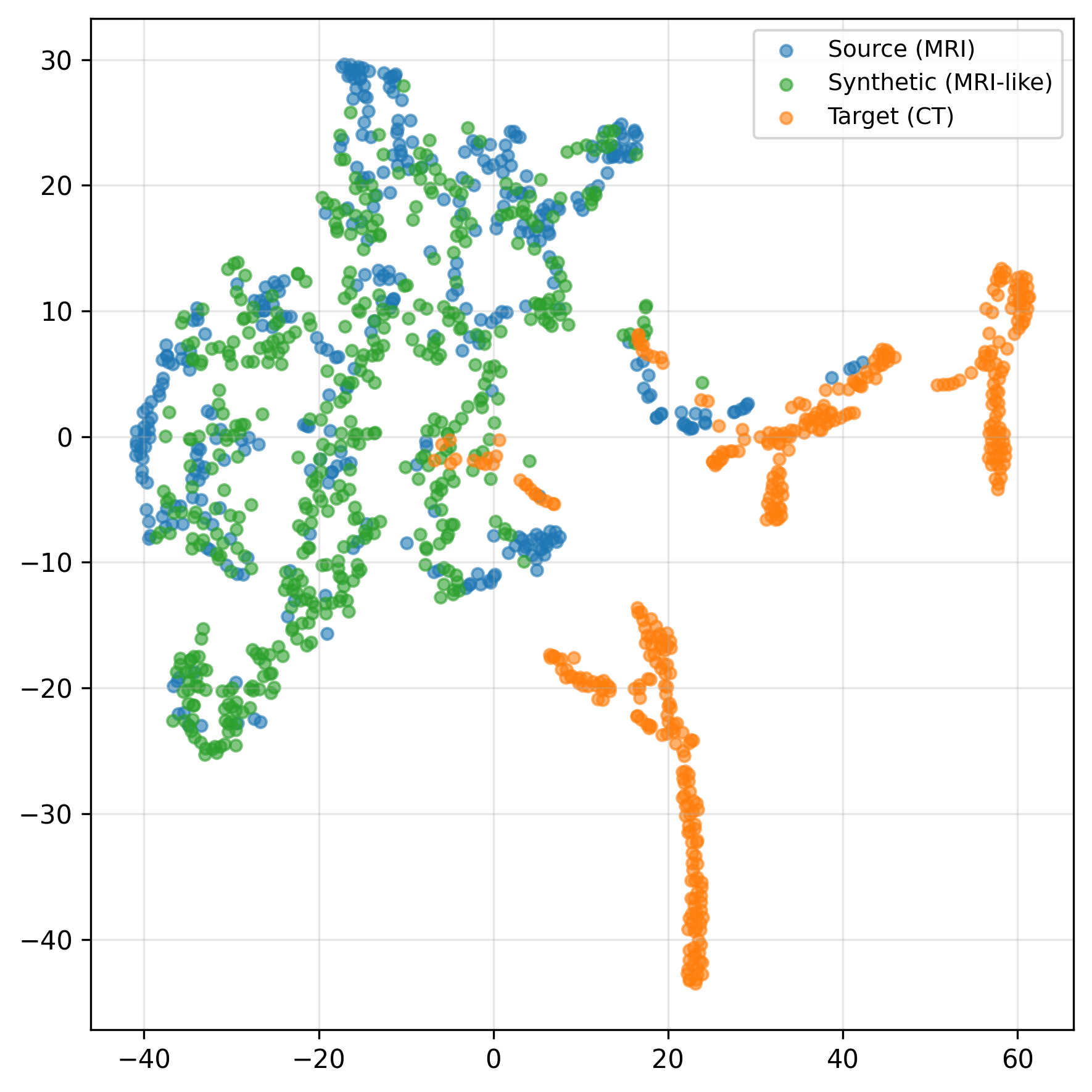}
        \end{minipage} \\[3pt]
        \textbf{(a)} & \textbf{(b)} & \textbf{(c)}
    \end{tabular}
    \caption{\textbf{Symmetrical Flow Matching for medical tasks.} SymmFlow
    learns a joint vector field from Gaussian noise to images ($X$) and masks
    ($Y$), enabling conditional segmentation ($Y|X$) and synthesis ($X|Y$).
    \textbf{(a)} Abdomen and cardiac examples produced by our source-trained model.
    \textbf{(b)} t-SNE shows overlapping mask features but separated image features.
    \textbf{(c)} Synthetic replay bridges the separated image-feature clusters.}
    \label{fig:symmflow_medical}
\end{figure}

\section{Method}
\label{sec:method}

\subsubsection{Problem Formulation.}
We consider a source domain $\mathcal{D}_s = \{(x_s^i, y_s^i)\}$ consisting of
paired images and segmentation masks, and a target domain $\mathcal{D}_t =
\{x_t^j\}$ containing only unlabeled images. We first train a SymmFlow model,
denoted as $\mathcal{M}_S$, on the source domain $\mathcal{D}_s$. In the
Source-Free Unsupervised Domain Adaptation (SFUDA) setting, we aim to adapt
$\mathcal{M}_S$ to the target domain $\mathcal{D}_t$ to maximize segmentation
performance, with the strict constraint that $\mathcal{D}_s$ is not accessible
during adaptation due to privacy or storage limitations.

\begin{figure}[t]
    \centering
    \includegraphics[scale=0.13]{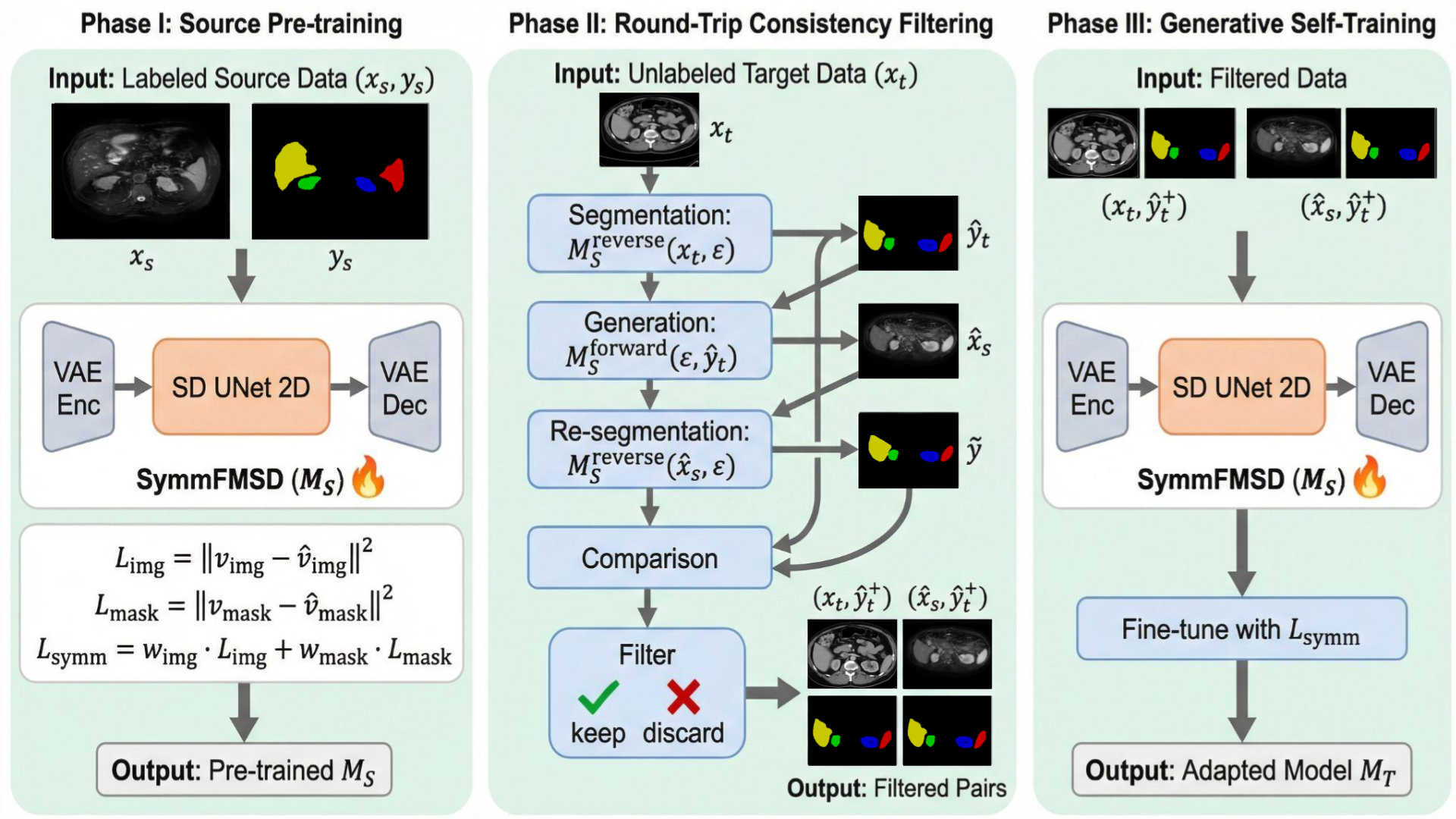}
    \caption{\textbf{SymmAdapt pipeline.} \textbf{Phase~I:} SymmFlow is trained
    on labeled source data, learning a bidirectional flow between images and
    masks. \textbf{Phase~II:} At adaptation time (no source data), target images
    are segmented and the resulting pseudo-masks are used to generate synthetic
    source-like images; a round-trip consistency filter retains only
    cycle-consistent pairs. \textbf{Phase~III:} The model is fine-tuned on a dual
    stream of real target and synthetic source-like pairs, anchoring source
    semantics while adapting to target appearance.}
    \label{fig:pipeline}
\end{figure}

\noindent\textbf{Symmetrical Flow Matching.}
Let $x$ and $y$ denote an image and its mask. Given $t \sim \mathcal{U}[0, 1]$
and noise $\xi_x, \xi_y \sim \mathcal{N}(0, I)$, SymmFlow defines
opposite-direction linear optimal transport paths and their velocity fields:
\begin{align}
x_t &= (1 - t)\xi_x + t x, & y_t &= (1 - t)y + t \xi_y
\label{eq:symmflow_interpolation}\\
v_x &= x - \xi_x, & v_y &= \xi_y - y
\label{eq:symmflow_velocity}
\end{align}
A network $v_\theta(x_t, y_t, t)$ predicts this joint velocity via MSE:
\begin{equation}
\mathcal{L}(\theta) = \mathbb{E}_{(x, y, t)} \left[ \| v_\theta(x_t, y_t, t) - v \|^2 \right]
\label{eq:symmflow_loss}
\end{equation}
Segmentation is obtained by integrating the mask ODE backward from $y_1 \sim
\mathcal{N}(0, I)$ ($t{=}1$) to the mask manifold ($t{=}0$), conditioned on the
image $x$:
\begin{equation}
\hat{y} = y_1 + \int_{1}^{0} v_{\theta}^{(y)}(x_t, y_t, t) \, dt
\label{eq:symmflow_inference}
\end{equation}
Fig.~\ref{fig:symmflow_medical} shows source-model examples and MRI$\to$CT
t-SNE views. Panel~(b) shows overlapping source-target mask features but
separated image features, indicating that anatomy transfers better than
modality-specific texture. Panel~(c) shows synthetic replay between the image
clusters, illustrating its role as a bridge during adaptation.
Figure~\ref{fig:pipeline} presents our SymmAdapt pipeline, detailed below.

\noindent\textbf{Phase I: Source Pre-training.} We train $\mathcal{M}_S$ on
labeled source data $\mathcal{D}_s$ following~\cite{symmflow}. Segmentation masks
are represented as RGB images, where each semantic class is assigned a unique
predefined RGB code. To prevent mode collapse from discrete pixel values during
flow matching, we apply \emph{label dequantization}: uniform noise $\epsilon \sim
\mathcal{U}(-\beta, +\beta)$ is added to the RGB mask values, converting them into
a continuous distribution suitable for stable flow training. Both dequantized
masks and images are encoded through the same VAE into a shared latent space. At
inference, the discrete segmentation map is recovered by assigning each pixel the
class whose predefined RGB code is closest to the predicted value.

\noindent\textbf{Phase II: Round-Trip Consistency Filtering.} Standard confidence
thresholding is unreliable under severe domain shift, so we retain only
pseudo-labels that survive a generative round-trip test. For a target image
$x_t$, we first obtain a candidate pseudo-mask $\hat{y}_t = \text{Segment}(x_t
\mid \mathcal{M}_S)$, then exploit SymmFlow's bidirectional nature to generate a
synthetic source-style image $\hat{x}_s = \text{Generate}(\hat{y}_t \mid
\mathcal{M}_S)$. If $\hat{y}_t$ captures valid anatomical semantics, $\hat{x}_s$
should be structurally coherent, and re-segmenting it should yield a mask
$\tilde{y}$ close to $\hat{y}_t$. We therefore retain only samples where
$\text{Dice}(\hat{y}_t, \tilde{y})$ exceeds both a global mean threshold and a
per-organ minimum, discarding predictions that fail this cycle-consistency check.

\noindent\textbf{Phase III: Generative Self-Training (GST).} We fine-tune
$\mathcal{M}_S$ on a dual-stream mixed batch. The \textit{target stream} $(x_t,
\hat{y}_t)$ uses filtered real target images with their pseudo-masks, aligning
the model's latent features with target-specific texture--shape correlations
(e.g., mapping dark MRI regions to liver shapes previously seen as bright in CT).
The \textit{synthetic stream} $(\hat{x}_s, \hat{y}_t)$ uses the source-like
images generated in Phase~II as a generative replay buffer that rehearses the
source manifold, providing anatomical anchoring that prevents the model from
drifting into degenerate solutions under large domain shifts. The objective is a
weighted decomposition of the joint MSE loss (Eq.~\ref{eq:symmflow_loss}):
\begin{equation}
    \mathcal{L}_{GST} = w_{\text{img}} \underbrace{\mathbb{E} [ \| v_{\theta}^{(x)} - v_x \|^2 ]}_{\mathcal{L}_{\text{img}}} + (1-w_{\text{img}}) \underbrace{\mathbb{E} [ \| v_{\theta}^{(y)} - v_y \|^2 ]}_{\mathcal{L}_{\text{mask}}}
\end{equation}
where $v_{\theta}^{(x)}$ and $v_{\theta}^{(y)}$ are the image and mask velocity
predictions. This continuous reinforcement of source semantic priors, combined
with target appearance alignment, ensures stable adaptation.

\begin{table}[tb]
    \renewcommand{\arraystretch}{1.0}
    \centering
    \caption{Quantitative evaluation results on the abdominal multi-organ datasets.}
    \label{tab:abdomen} \resizebox{1.0\textwidth}{!}{%
        \setlength{\tabcolsep}{3pt}
        \begin{tabular}{l c | c c c c c | c c c c c}
            \toprule \multicolumn{1}{c}{\multirow{2}{*}{\textbf{Method (MRI$\rightarrow$CT)}}}    & \multirow{2}{*}{\textbf{SFUDA}} & \multicolumn{5}{c|}{\textbf{Dice (\% $\uparrow$)}} & \multicolumn{5}{c}{\textbf{ASSD (mm $\downarrow$)}}                                                                                                                                                           \\
            \cmidrule(lr){3-7} \cmidrule(lr){8-12} \multicolumn{2}{c|}{}                          & \textbf{Liver}                  & \textbf{R.kidney}                                  & \textbf{L.kidney}                                   & \textbf{Spleen}  & \textbf{Average} & \textbf{Liver}   & \textbf{R.kidney} & \textbf{L.kidney} & \textbf{Spleen}  & \textbf{Average}                    \\
            \midrule
            Target Supervised                                                                     & --                              & 95.4                                               & 89.6                                                & 90.3             & 94.2             & 92.4             & 0.64              & 1.39              & 0.58             & 0.52             & 0.78             \\
            \hline
            SIFA \cite{sifa}                                                                      & \xmark                          & 88.0                                               & 83.3                                                & 80.9             & 82.6             & 83.7             & 1.2               & 1.0               & 1.5              & 1.6              & 1.3              \\
            Diffusion-based DA \cite{Ji_Diffusionbased_MICCAI2024}                                & \xmark                          & 89.0                                               & 85.6                                                & 85.6             & 85.8             & 86.5             & 1.5               & 1.3               & 1.2              & 1.2              & 1.3              \\
            C3R \cite{c3r}                                                                        & \xmark                          & 91.9                                               & 84.3                                                & 82.6             & 85.0             & 86.0             & 0.93              & 0.98              & 1.10             & 1.39             & 1.10             \\
            \hline
            DPL \cite{dpl}                                                                        & \cmark                          & 39.0                                               & 35.3                                                & 58.6             & 42.4             & 43.9             & 6.36              & 13.09             & 7.94             & 10.10            & 9.38             \\
            AdaMI \cite{adami}                                                                    & \cmark                          & 72.8                                               & 57.0                                                & 63.5             & 47.1             & 60.1             & 5.51              & 11.54             & 10.7             & 10.78            & 9.64             \\
            UPL \cite{upl}                                                                        & \cmark                          & 78.1                                               & 61.0                                                & 68.2             & 38.9             & 61.5             & 4.21              & 6.80              & 6.95             & 9.31             & 6.82             \\
            FVP \cite{fvp}                                                                        & \cmark                          & 87.8                                               & 64.7                                                & 73.2             & 68.3             & 73.5             & 3.63              & 2.58              & 3.10             & \underline{2.34} & 2.91             \\
            ProtoContra \cite{protocontra}                                                        & \cmark                          & 85.4                                               & 75.5                                                & 68.0             & 68.9             & 74.4             & 3.22              & 5.91              & 6.17             & 4.18             & 4.87             \\
            SFS \cite{sfs}                                                                        & \cmark                          & 88.3                                               & 73.7                                                & 80.7             & \underline{81.6} & 81.1             & 2.4               & 4.1               & 3.5              & 2.7              & 3.2              \\
            DFG \cite{dfg}                                                                        & \cmark                          & \underline{90.9}                                   & \textbf{88.9}                                       & \underline{82.7} & 77.3             & \underline{84.9} & \textbf{1.29}     & \textbf{0.66}     & \textbf{1.74}    & 4.28             & \textbf{1.99}    \\
            ProtoContra$^\dagger$ \cite{protocontra}                                              & \cmark                          & 84.3                                               & 73.0                                                & 64.5             & 62.7             & 71.1             & 4.46              & 7.48              & 11.02            & 10.89            & 8.46             \\
            DFG$^\dagger$ \cite{dfg}                                                              & \cmark                          & 91.0                                               & 80.3                                                & 83.4             & 74.6             & 82.3             & \underline{1.44}  & \underline{1.02}  & \underline{2.03} & 4.86             & 2.34             \\
            \hline
            SymmAdapt (Ours)                                                                      & \cmark                          & \textbf{92.8}                                      & \underline{82.8}                                    & \textbf{84.2}    & \textbf{88.8}    & \textbf{87.2}    & 2.84              & 1.81              & 2.60             & \textbf{1.02}    & \underline{2.07} \\
            \bottomrule \multicolumn{1}{c}{\multirow{2}{*}{\textbf{Method (CT$\rightarrow$MRI)}}} & \multirow{2}{*}{\textbf{SFUDA}} & \multicolumn{5}{c|}{\textbf{Dice (\% $\uparrow$)}} & \multicolumn{5}{c}{\textbf{ASSD (mm $\downarrow$)}}                                                                                                                                                           \\
            \cmidrule(lr){3-7} \cmidrule(lr){8-12} \multicolumn{2}{c|}{}                          & \textbf{Liver}                  & \textbf{R.kidney}                                  & \textbf{L.kidney}                                   & \textbf{Spleen}  & \textbf{Average} & \textbf{Liver}   & \textbf{R.kidney} & \textbf{L.kidney} & \textbf{Spleen}  & \textbf{Average}                    \\
            \midrule
            Target Supervised                                                                     & --                              & 91.9                                               & 91.3                                                & 90.7             & 90.5             & 91.1             & 0.52              & 0.45              & 0.35             & 0.93             & 0.56             \\
            \hline
            SIFA \cite{sifa}                                                                      & \xmark                          & 90.0                                               & 89.1                                                & 80.2             & 82.3             & 85.4             & 1.5               & 0.6               & 1.5              & 2.4              & 1.5              \\
            Diffusion-based DA \cite{Ji_Diffusionbased_MICCAI2024}                                & \xmark                          & 84.4                                               & 90.3                                                & 92.1             & 86.6             & 88.3             & 1.5               & 0.5               & 0.5              & 0.6              & 0.8              \\
            C3R \cite{c3r}                                                                        & \xmark                          & 91.6                                               & 91.2                                                & 92.0             & 90.0             & 91.2             & 1.08              & 0.47              & 0.64             & 0.91             & 0.77             \\
            \hline
            DPL \cite{dpl}                                                                        & \cmark                          & 63.1                                               & 65.3                                                & 57.3             & 38.3             & 56.0             & 2.86              & 1.14              & 1.40             & 2.61             & 2.00             \\
            AdaMI \cite{adami}                                                                    & \cmark                          & 54.8                                               & 77.2                                                & 64.4             & 47.6             & 61.0             & 4.90              & 5.75              & 4.02             & 5.87             & 5.14             \\
            UPL \cite{upl}                                                                        & \cmark                          & 56.9                                               & 55.6                                                & 59.9             & 55.3             & 56.9             & 3.86              & 1.47              & 1.58             & 4.77             & 2.92             \\
            FVP \cite{fvp}                                                                        & \cmark                          & 64.8                                               & 87.6                                                & 80.3             & 60.5             & 73.3             & 4.48              & 2.10              & 1.54             & 6.15             & 3.57             \\
            ProtoContra \cite{protocontra}                                                        & \cmark                          & 71.6                                               & 83.8                                                & 81.6             & 84.9             & 80.5             & 2.89              & 3.36              & 4.73             & 1.09             & 3.02             \\
            SFS \cite{sfs}                                                                        & \cmark                          & \underline{86.3}                                   & \underline{88.0}                                    & \underline{85.1} & 74.9             & 83.5             & 4.5               & 1.6               & 2.2              & 18.2             & 6.6              \\
            DFG \cite{dfg}                                                                        & \cmark                          & 83.6                                               & \textbf{92.5}                                       & 79.6             & \underline{85.1} & \underline{85.2} & \underline{1.82}  & \textbf{0.30}     & \underline{0.87} & \underline{1.02} & \underline{1.00} \\
            \hline
            SymmAdapt (Ours)                                                                      & \cmark                          & \textbf{87.5}                                      & 87.8                                                & \textbf{85.9}    & \textbf{90.7}    & \textbf{88.0}    & \textbf{1.34}     & \underline{0.76}  & \textbf{0.61}    & \textbf{0.53}    & \textbf{0.81}    \\
            \bottomrule
        \end{tabular}
    }
\end{table}

\section{Experiments}
\label{sec:experiments}

We evaluate on three standard cross-modality UDA benchmarks (abdominal, cardiac,
prostate), strictly following established data splits and preprocessing from
prior work~\cite{dfg,sifa,adami} to isolate the algorithmic contribution.
\noindent\textbf{Abdominal Multi-Organ (CT $\leftrightarrow$ MRI).} BTCV (30 CT
volumes)~\cite{btcv_miccai2015} and CHAOS (20 T2-SPIR MRI volumes)~\cite{chaos};
four organs (Liver, Spleen, Right/Left Kidney); 80:20 split per modality.
\noindent\textbf{Cardiac Substructures (MRI $\rightarrow$ CT).} MMWHS~\cite{mmwhs}
with 20 unpaired CT and 20 MRI volumes; four substructures (AA, LAC, LVC, MYO);
16/4 train/test split per modality following~\cite{sifa}.
\noindent\textbf{Prostate (Cross-Site MRI).} RUNMC (30, 3T Siemens) $\to$ BMC
(30, 1.5T Philips) from NCI-ISBI 2013~\cite{nci-isbi} with a 20/10
split~\cite{adami}; and QUBIQ (55 inherently 2D cases)~\cite{qubiq} using the
official split~\cite{dfg}.

\subsection{Implementation Details}
\label{sec:impl}
We implement $\mathcal{M}_S$ using the SymmFMSD architecture~\cite{symmflow} with
the pre-trained Stable Diffusion 2.1 VAE and U-Net~\cite{Rombach_2022_CVPR},
interpolating slices to $512 \times 512$ before encoding. Inference uses Euler
ODE integration with 5 steps at 5.6\,s per abdomen volume.
Source pre-training uses AdamW with cosine annealing for 100
epochs; label dequantization noise $\beta = 6.0$; and loss weights $w_{\text{img}}
= 0.1$, $w_{\text{mask}} = 0.9$. During adaptation, the round-trip filter requires
mean Dice $\ge 0.85$ and per-organ Dice $\ge 0.70$; GST fine-tunes with AdamW at
$\text{lr}=10^{-5}$ for 5--10 epochs. We train at slice level and evaluate at
volume level using Dice and ASSD on a single NVIDIA RTX 6000 (48GB).

\subsection{Results}
\label{sec:results}

\begin{figure}[tb]
    \includegraphics[width=\linewidth]{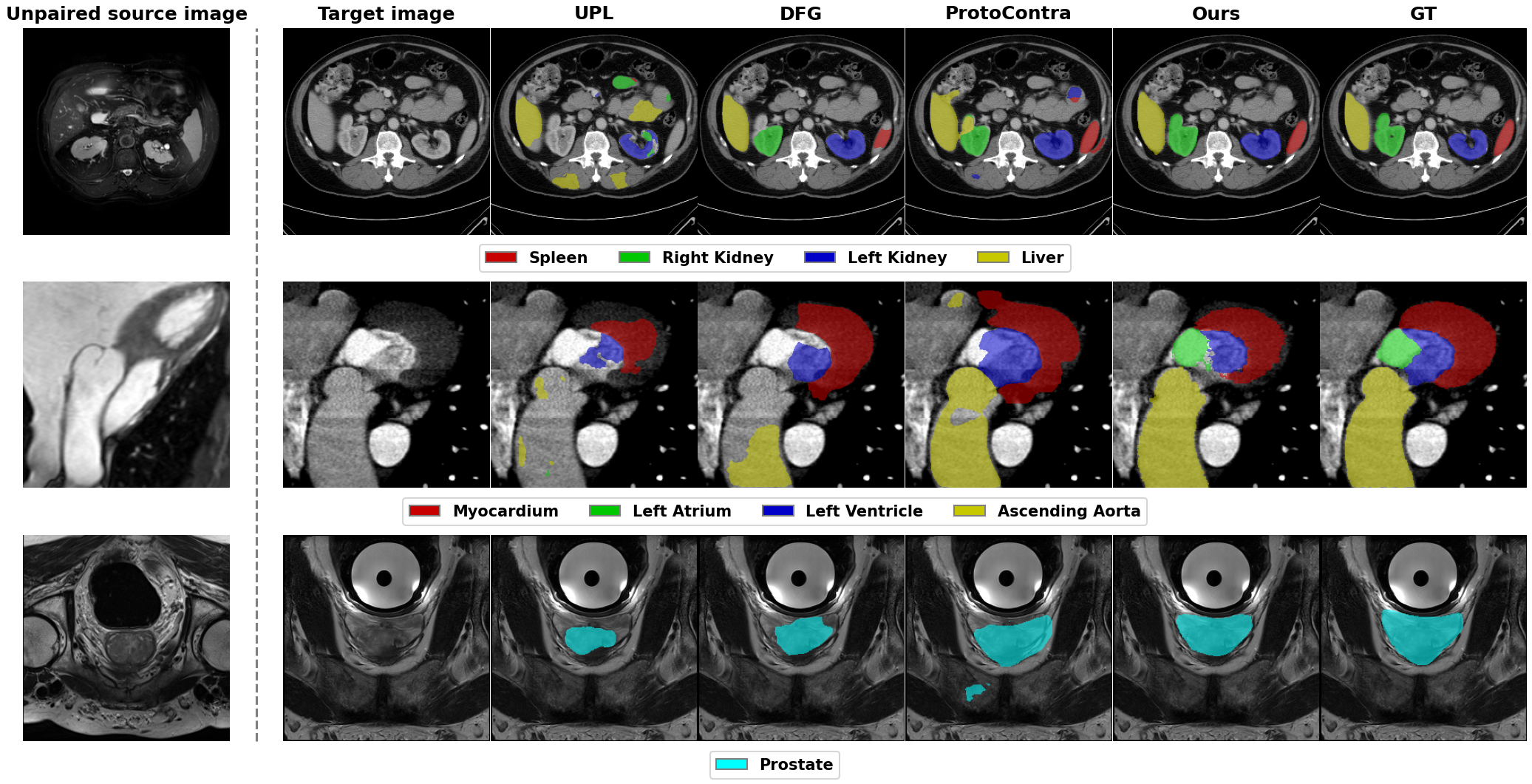}
    \caption{Qualitative results across three benchmarks (rows: abdomen MRI$\to$CT, cardiac MRI$\to$CT, prostate RUNMC$\to$QUBIQ). SymmAdapt yields smoother boundaries and fewer false positives, closer to the ground truth. The first column shows a representative source-domain image.}
    \label{fig:qualitative}
\end{figure}

We compare SymmAdapt against UDA and SFUDA methods. Baselines are from original
publications or from DFG~\cite{dfg}; for MRI$\to$CT and prostate benchmarks
we also re-implement DFG~\cite{dfg}, ProtoContra~\cite{protocontra}, and
UPL~\cite{upl} using identical splits and original hyperparameters (rows marked $^\dagger$). Among SFUDA methods, \textbf{best}
and \underline{second-best} results are highlighted;
``Target Supervised'' denotes the upper bound achieved by our own model when trained with fully supervised target labels (oracle).

\noindent\textbf{Abdominal Multi-Organ (Table~\ref{tab:abdomen}):} SymmAdapt
achieves 87.2\% and 88.0\% average Dice for MRI$\to$CT and CT$\to$MRI,
outperforming the best SFUDA baseline DFG by +2.8 and +1.7 points, while remaining competitive with non-source-free methods.
\textbf{Cardiac MMWHS (Table~\ref{tab:mmwhs}):} our method yields 87.2\% Dice for
MRI$\to$CT, surpassing the best SFUDA baseline ProtoContra (84.5\%)~\cite{protocontra}
and non-source-free C3R (86.1\%)~\cite{c3r}.
\begin{table}[tb]
    \renewcommand{\arraystretch}{1.0}
    \centering
    \caption{Quantitative evaluation results on the MMWHS cardiac dataset.}
    \label{tab:mmwhs} \resizebox{1.0\textwidth}{!}{%
        \setlength{\tabcolsep}{3pt}
        \begin{tabular}{l c | c c c c c | c c c c c}
            \toprule \multicolumn{1}{c}{\multirow{2}{*}{\textbf{Method (MRI$\rightarrow$CT)}}} & \multirow{2}{*}{\textbf{SFUDA}} & \multicolumn{5}{c|}{\textbf{Dice (\% $\uparrow$)}} & \multicolumn{5}{c}{\textbf{ASSD (mm $\downarrow$)}}                                                                                                                                                        \\
            \cmidrule(lr){3-7} \cmidrule(lr){8-12} \multicolumn{2}{c|}{}                       & \textbf{AA}                     & \textbf{LAC}                                       & \textbf{LVC}                                        & \textbf{MYO}     & \textbf{Average} & \textbf{AA}      & \textbf{LAC}     & \textbf{LVC}     & \textbf{MYO}    & \textbf{Average}                    \\
            \midrule
            Target Supervised                                                                  & --                              & 94.8                                               & 92.3                                                & 92.1             & 89.8             & 92.2             & 1.96             & 4.32             & 2.14            & 1.67             & 2.52             \\
            \hline
            SIFA \cite{sifa}                                                                   & \xmark                          & 81.3                                               & 79.5                                                & 73.8             & 61.6             & 74.1             & 7.9              & 6.2              & 5.5             & 8.5              & 7.0              \\
            C3R \cite{c3r}                                                                     & \xmark                          & 90.2                                               & 90.2                                                & 86.6             & 77.5             & 86.1             & 4.72             & 2.81             & 2.58            & 3.07             & 3.30             \\
            GenericSSL \cite{genericssl}                                                       & \xmark                          & 93.2                                               & 89.5                                                & 91.7             & 86.2             & 90.1             & --               & --               & --              & --               & 1.7              \\
            \hline
            AdaMI \cite{adami}                                                                 & \cmark                          & 83.1                                               & 78.2                                                & 74.5             & 66.8             & 75.7             & 5.6              & 4.2              & 5.7             & 6.9              & 5.6              \\
            DPL \cite{dpl}                                                                     & \cmark                          & \underline{91.0}                                   & 69.4                                                & 78.2             & 65.2             & 76.0             & 8.56             & 9.00             & 6.11            & 4.85             & 7.13             \\
            FVP \cite{fvp}                                                                     & \cmark                          & 85.6                                               & 71.9                                                & 79.5             & 64.0             & 75.3             & 9.01             & 9.00             & 4.37            & \underline{3.52} & 6.48             \\
            SFS \cite{sfs}                                                                     & \cmark                          & 88.0                                               & 83.7                                                & 81.0             & 72.5             & 81.3             & 6.3              & 7.2              & \underline{4.7} & 6.1              & 6.1              \\
            DFG$^\dagger$ \cite{dfg}                                                           & \cmark                          & 83.4                                               & 86.8                                                & 76.5             & 67.1             & 78.4             & \underline{4.65} & \underline{4.06} & 5.32            & 4.41             & \underline{4.61} \\
            ProtoContra$^\dagger$ \cite{protocontra}                                           & \cmark                          & 87.7                                               & \underline{87.5}                                    & \underline{86.6} & \underline{76.2} & \underline{84.5} & 9.90             & 6.00             & 5.29            & 6.90             & 7.03             \\
            \hline
            SymmAdapt (Ours)                                                                   & \cmark                          & \textbf{93.2}                                      & \textbf{91.4}                                       & \textbf{89.2}    & \textbf{74.9}    & \textbf{87.2}    & \textbf{4.46}    & \textbf{3.15}    & \textbf{2.22}   & \textbf{3.33}    & \textbf{3.29}    \\
            \bottomrule
        \end{tabular}
    }
\end{table}

\textbf{Prostate (Table~\ref{tab:prostate_adapt}):} SymmAdapt is best on both
RUNMC$\to$QUBIQ and RUNMC$\to$BMC, exceeding DFG~\cite{dfg} and
AdaMI~\cite{adami} on Dice and ASSD.
\begin{table}[tb]
    \renewcommand{\arraystretch}{1.0}
    \centering
    \caption{Quantitative evaluation results on the prostate task.}
    \label{tab:prostate_adapt} \resizebox{1.0\textwidth}{!}{%
    \setlength{\tabcolsep}{3pt}
    \begin{tabular}{l c | c c | c c}
        \toprule \multicolumn{1}{c}{\multirow{2}{*}{\textbf{Method}}} & \multirow{2}{*}{\textbf{SFUDA}} & \multicolumn{2}{c|}{\textbf{RUNMC $\rightarrow$ QUBIQ}} & \multicolumn{2}{c}{\textbf{RUNMC $\rightarrow$ BMC}} \\
        \cmidrule(lr){3-4} \cmidrule(lr){5-6} \multicolumn{2}{c|}{}   & \textbf{Dice (\% $\uparrow$)}   & \textbf{ASSD (mm $\downarrow$)}                         & \textbf{Dice (\% $\uparrow$)}                       & \textbf{ASSD (mm $\downarrow$)} \\
        \midrule
        Target Supervised                                   & --                              & 96.5                                                    & 0.09                                                & 89.2                           & 0.77             \\
        \hline
        DPL \cite{dpl}                                                & \cmark                          & 75.5                                                    & 8.33                                                & --                             & --               \\
        UPL \cite{upl}                                                & \cmark                          & 73.6                                                    & 16.55                                               & --                             & --               \\
        ProtoContra \cite{protocontra}                                & \cmark                          & 86.6                                                    & 5.58                                                & --                             & --               \\
        AdaMI \cite{adami}                                            & \cmark                          & 81.0                                                    & 8.46                                                & \underline{79.5}               & \underline{3.92} \\
        DFG \cite{dfg}                                                & \cmark                          & \underline{93.3}                                        & \underline{2.18}                                    & --                             & --               \\
        UPL$^\dagger$ \cite{upl}                                      & \cmark                          & 74.2                                                    & 1.71                                                & --                             & --               \\
        ProtoContra$^\dagger$ \cite{protocontra}                      & \cmark                          & 88.8                                                    & 1.91                                                & --                             & --               \\
        DFG$^\dagger$ \cite{dfg}                                      & \cmark                          & 83.7                                                    & 0.81                                                & --                             & --               \\
        \hline
        SymmAdapt (Ours)                                              & \cmark                          & \textbf{94.3}                                          & \textbf{0.30}                                       & \textbf{85.0}                  & \textbf{1.02}    \\
        \bottomrule
    \end{tabular}
    }
\end{table}

\noindent\textbf{Ablation Study (Table~\ref{tab:ablation_filtering}):} We ablate
the adaptation components and source-training design on abdomen MRI$\to$CT.
The component ablation shows that source-only SymmFlow is already strong
(84.4\%), target-only fine-tuning drifts (68.7\%), synthetic replay restores the
source anchor (84.1\%), and round-trip filtering yields the full 87.2\%.
Panel~(b) ablates the source training formulation by training three models on
the same SD-UNet backbone. Direct MSE scores 92.8\% on MRI but only 77.0\% on
CT; SemFlow~\cite{semflow}, which learns a direct image$\leftrightarrow$mask
flow without intermediate Gaussian initialization, collapses to 36.1\%. SymmFlow retains
84.4\%, supporting our hypothesis that initializing from a fixed Gaussian origin when using flow matching
is key to robustness under domain shift.

\begin{center}
    \refstepcounter{table}\label{tab:ablation_filtering}
    \renewcommand{\arraystretch}{0.95}
    \footnotesize
    \textbf{Table~\thetable. Ablation study on abdomen MRI$\rightarrow$CT.}
    \textbf{(a)} Adaptation components.
    \textbf{(b)} Source training design.
    Dice (\%).\\[3pt]
    \begin{minipage}[t]{0.48\textwidth}
        \centering
        \setlength{\tabcolsep}{2pt}
        \resizebox{\linewidth}{!}{%
        \begin{tabular}{c c c | c}
            \toprule
            \textbf{Round-trip} & \textbf{Target} & \textbf{Synth.} &
            \textbf{Dice} \\
            \midrule
            \xmark & \xmark & \xmark & 84.4 \\
            \cmark & \cmark & \xmark & 68.7 \\
            \xmark & \cmark & \cmark & 84.1 \\
            \cmark & \cmark & \cmark & \textbf{87.2} \\
            \bottomrule
        \end{tabular}}
        \\[2pt] \footnotesize (a)
    \end{minipage}
    \hfill
    \begin{minipage}[t]{0.48\textwidth}
        \centering
        \setlength{\tabcolsep}{2pt}
        \resizebox{\linewidth}{!}{%
        \begin{tabular}{l | c c}
            \toprule
            \textbf{Source model} & \textbf{MRI} & \textbf{MRI$\to$CT} \\
            \midrule
            Direct MSE              & \textbf{92.8} & 77.0 \\
            SemFlow~\cite{semflow}  & 85.7          & 36.1 \\
            SymmFlow (Ours)         & 91.4          & \textbf{84.4} \\
            \bottomrule
        \end{tabular}}
        \\[2pt] \footnotesize (b)
    \end{minipage}
\end{center}

\subsection{Analysis: Efficiency and Stability}
\label{sec:analysis}
\noindent\textbf{Inference cost.} Although flow-matching inference solves an ODE,
very few steps suffice: 1, 5, 10, 25, and 50 Euler steps give Dice of 87.0,
87.2, 86.0, 86.0, and 86.0, respectively, with 5 steps giving the best ASSD
(2.07\,mm) and 5.6\,s per volume.

\noindent\textbf{Pseudo-label quality and retention.} Round-trip consistency
filtering retains the vast majority of pseudo-labels: at 5 ODE steps, 1142 of
1356 slices (84.2\%) pass the filter with a mean round-trip Dice of 0.903,
indicating high-quality self-supervision. We further train GST with
$N\in\{1,5,25\}$ ODE steps and with loose, paper, and strict thresholds. All
five runs converge within $<$0.5 Dice points of each other, confirming that the
filtering threshold is not a sensitive hyperparameter.

\noindent\textbf{Stochastic stability.} SymmAdapt starts every inference from a
random Gaussian sample $z_0\!\sim\!\mathcal{N}(0,I)$, so predictions vary with
the noise seed. We run 10 inferences per test volume, each with a different
seed. The 3D-Dice standard deviation is clinically negligible:
$\le 0.014$ per organ on abdomen (avg.\ 0.011) and $\le 0.006$ on MMWHS
(avg.\ 0.003).

\section{Conclusion}
\label{sec:conclusion}
We introduced SymmAdapt, a Source-Free Domain Adaptation framework powered by
Symmetrical Flow Matching. Our results show that Gaussian-origin inference
preserves anatomical priors under severe appearance shifts, synthetic replay
provides source-like anchors without source data, and round-trip filtering
stabilizes self-training on unlabeled target images. Across abdominal, cardiac,
and prostate benchmarks, SymmAdapt achieves strong SFUDA performance and remains
competitive with non-source-free methods. Models such as the one presented here
can support clinic-focused segmentation when source data cannot be shared and
domain shifts hinder deployment. A current limitation is that adaptation
requires a flow-matching source model rather than an arbitrary pretrained
segmenter. Future work will distill existing segmenters into flow models and
explore Test-Time Training for single-instance adaptation.

\bibliographystyle{splncs04}
\bibliography{references}
\end{document}